\documentclass{article}
\usepackage[final]{nips_2017}
\usepackage[utf8]{inputenc} 
\usepackage[T1]{fontenc}    
\usepackage{hyperref}       
\usepackage{url}            
\usepackage{booktabs}       
\usepackage{amsfonts}       
\usepackage{nicefrac}       
\usepackage{microtype}      
\usepackage{graphicx}
\usepackage{afterpage}

\usepackage{setspace}  

\usepackage{chngpage}

\usepackage{booktabs}

\usepackage{caption}
\usepackage{subcaption}

\title{Exploring Feature Reuse in DenseNet Architectures}

\author{
  Andy Hess\\
  Department of Computer Science\\
  University of Alberta\\
  \texttt{athess@ualberta.ca} \\
}

\begin{document}




\maketitle

\doublespacing

\begin{abstract}
Densely Connected Convolutional Networks (DenseNets) [1] have been shown to achieve state-of-the-art results on image classification tasks while using fewer parameters and computation than competing methods. Since each layer in this architecture has full access to the feature maps of all previous layers, the network is freed from the burden of having to relearn previously useful features, thus alleviating issues with vanishing gradients. In this work we explore the question: To what extent is it necessary to connect to \emph{all} previous layers in order to reap the benefits of feature reuse? To this end, we introduce the notion of \emph{local dense connectivity} and present evidence that less connectivity, allowing for increased growth rate at a fixed network capacity, can achieve a more efficient reuse of features and lead to higher accuracy in dense architectures.

\end{abstract}

\section{Introduction}
Deep networks have have been getting deeper in recent years [2,3,4] and with increased depth, challenges such as vanishing gradient and other issues can arise. To combat these issues, architectures have been proposed that connect more distant layers directly to other layers. Generally referred to as \emph{identity connections} or \emph{skip connections}, methods include variations not only of the network topology but also the nature through which connectivity occurs.

Inspired by the gating mechanism found in LSTMs [5], layers in Highway Networks [6] learn to regulate information flow across local skip connections. Instead of learning to gate local connections, ResNets [4] perform a similar type of regulation but instead by learning residual functions. In this case, information is carried from previous layers through the addition operation. 

Other networks feature connectivity patterns that carry information directly across larger depths. Training deep ResNets with Stochastic Depth [7] does this implicitly during training; as subsets of layers are randomly dropped (for each mini-batch) and replaced with the identity function. By using Drop-Path training of Fractalnets [8], where connections from a fractal-inspired architecture are dropped during training, ResNet-level [4] performance is achieved but without using residual connections.

This trend towards increased connectivity and across greater depths culminated in the DenseNet architecture [1] wherein every layer is connected to every other layer.

In this work, we relax the fully-dense connectivity of DenseNet by introducing network architectures where dense connectivity within each dense block is limited to only $N$ previous layers. Since we parameterize these networks using this \emph{dense window size $N$}, we refer to these architectures as \emph{WinDenseNet-N}. Comparing to the baseline model DenseNet-40 [1] on CIFAR-10 [12], we show that limiting connectivity this way can greatly reduce the number of parameters (and thus training time) of these networks with only a small reduction in accuracy. More importantly, we provide evidence that WinDenseNets, at various window sizes, can utilize parameter capacity more effectively than DenseNets. In other words, for a fixed capacity, networks with lower dense connectivity and higher growth rate can outperform their fully-dense counterparts. Further, we provide insight into why this may be the case by visualizing feature reuse in these dense architectures.

\section{ Methods }

\begin{figure}[t]
\begin{center}
   \includegraphics[width=0.7\textwidth]{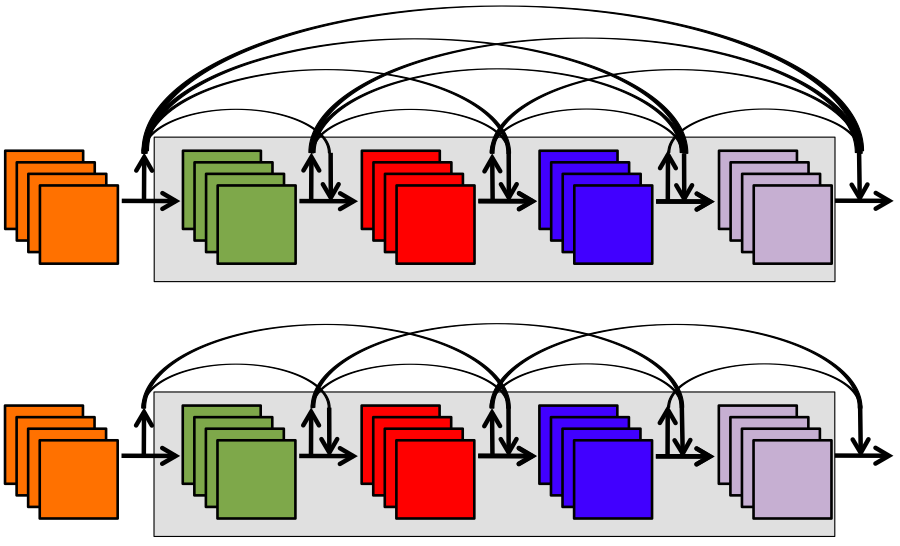}
   \caption{\textbf{Top}: Connectivity pattern of DenseNet [1] where every layer in a dense block (gray) has access to the concatenation of all previous feature maps. Note that original input maps (orange) are passed even across the entire dense block (largest arc). \textbf{Bottom}: Our proposed windowed dense connectivity pattern (value of $3$ shown). For a dense window value of N, each layer receives feature maps from at most, N preceding layers. Note, for a dense block with $L$ layers, connectivity with a dense window of size $L+1$ is equivalent to DenseNet [1].}
   \label{figureDenseConnect}
  \end{center}
\end{figure}

The DenseNet architecture [1] consists of a series of \emph{dense blocks} where each layer within a dense block is densely connected to \emph{all} preceding layers (one dense block shown at top of figure \ref{figureDenseConnect}). Notably, and in contrast to the additive connectivity of ResNets, information flow in DenseNets occurs using feature map concatenation. Since feature concatenation can only be performed with feature maps of the same size, dense blocks are separated by \emph{transition layers} wherein down-sampling occurs via pooling. Capacity is parameterized by the number of new convolutional feature maps generated at each layer: the growth rate $k$ of the network.

Our proposed architecture explores limiting the connectivity of a target layer within dense blocks to only $N$ previous (source) layers. For example, for a dense connectivity window size of $3$, each layer in a dense block takes as input, the concatenation of features maps from at most $3$ preceding source layers (bottom of figure \ref{figureDenseConnect}). For this example, note that for the 1st and 2nd layers, only $1$ and $2$ source layers respectively, are taken as input. Further, note that the (transition or final) layer that immediately follows this dense block also takes the $3$ prior layers as input.

For a dense block having $L$ layers, a dense connectivity window of size $L + 1$ is required for the layer immediately following a dense block to be able to reach back to include feature maps that first entered the dense block (orange maps in figure \ref{figureDenseConnect}) and, in this case, the network has the equivalent of full DenseNet connectivity. Since maps first entering dense blocks can often number in the hundreds, lowering dense connectivity even by one can lead to a dramatic drop in the number of trainable parameters. For a fixed capacity, the notion of limited dense connectivity explores the idea of dropping distant (potentially less useful) features in exchange for the improvement that can be gained by increasing network growth rate (number of filters/feature maps at each layer). In the following section we show how our proposed locally dense-connected networks can lead to a more efficient use of parameters and, subsequently, higher accuracy than DenseNet-40 given a fixed capacity.

\section{Experiments}
For our experiments we begin with the reference DenseNet-40 model from [1] with default dense block growth rate of $12$. This base architecture contains $3$ dense blocks each having $12$ densely connected layers each comprised of Batch Normalization, ReLU and 3x3 convolution. The total number of trainable parameters of this network is $1,019,722$. Neither bottleneck nor transition layer compression is used in our experiments. Training times are reported from training on an NVIDIA GeForce GTX 1080 Ti Graphics Card.

In all our experiments, unless otherwise noted, all training hyper-parameters were kept the same as the original paper. Specifically, SGD was used with a batch size of $64$, momentum of $0.9$, weight decay of $1e-4$ and dropout set to retain $80\%$. The learning rate schedule begins with $0.1$ then changes to $0.01$ at epoch $150$ and is finally lowered to $0.001$ at epoch $225$. Training on CIFAR-10 [12] ends after $300$ epochs and test results on the test portion are reported. We do not use data augmentation. For code, we expand upon the Tensorflow [9] port [10] of the original Torch implementation [11].

The original DenseNet paper [1] reported a no-augmentation DenseNet-40 result of $93.00\%$ accuracy on CIFAR-10. We achieved a comparable result of $92.65\%$. This small difference may be due to using TensorFlow instead of their original Torch implementation and/or other factors such as random initialization etc.

\subsection{Dense Window Connectivity}


\begin{figure}[t]
\begin{center}
   \includegraphics[width=0.9\textwidth]{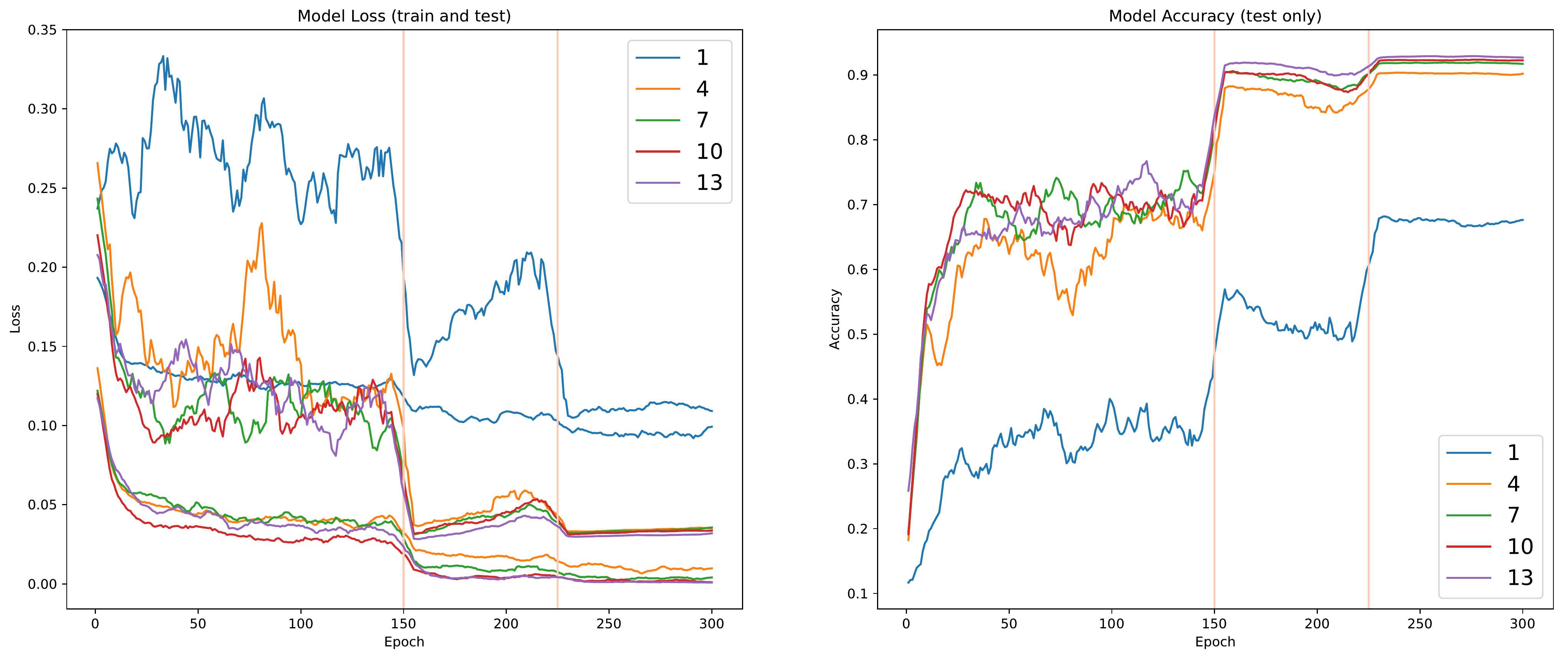}

   \caption{Training curves for cross entropy loss and model accuracy at dense connectivity window sizes of $1$, $4$, $7$, $10$ and $13$ showing that our proposed models, having far less dense connectivity than DenseNet [1], and far fewer parameters, attain almost the same level of accuracy (WinDenseNet-13 here is equivalent to DenseNet-40). Note that the model with a dense window size of $1$ struggles to learn (this corresponds to a traditional convnet with 1-connectivity). Note the loss graph on left has two curves for each window size (upper=test, lower=train). The orange vertical lines correspond to when reductions in learning rate occur. Graph values above are mean-smoothed over a range of $11$ epochs.}
   \label{figureTrainingCurves}
  \end{center}
\end{figure}

Leaving all other factors the same, we propose to modify the reference, $12$ layers per block, DenseNet architecture by varying the amount of local dense connectivity and measuring the affect on accuracy, training time and number of trainable network parameters. For these experiments we vary the dense window size from $1$ to $13$. A value of $1$ is similar to a traditional feed-forward convolutional architecture; where filters of any given layer convolve only over the previous layer's feature maps. A dense window value of $13$ corresponds to each layer within a dense block having convolutional access to all previous layers within the dense block. Note that a value of $13$ is necessary for the transition or final layer that follows the dense block to have access to that dense block's input maps (with a dense window size of only $12$, the subsequent transition layer can no longer reach the maps that were first input into the block).

\begin{figure}[t]
\begin{center}
   \includegraphics[width=1.0\textwidth]{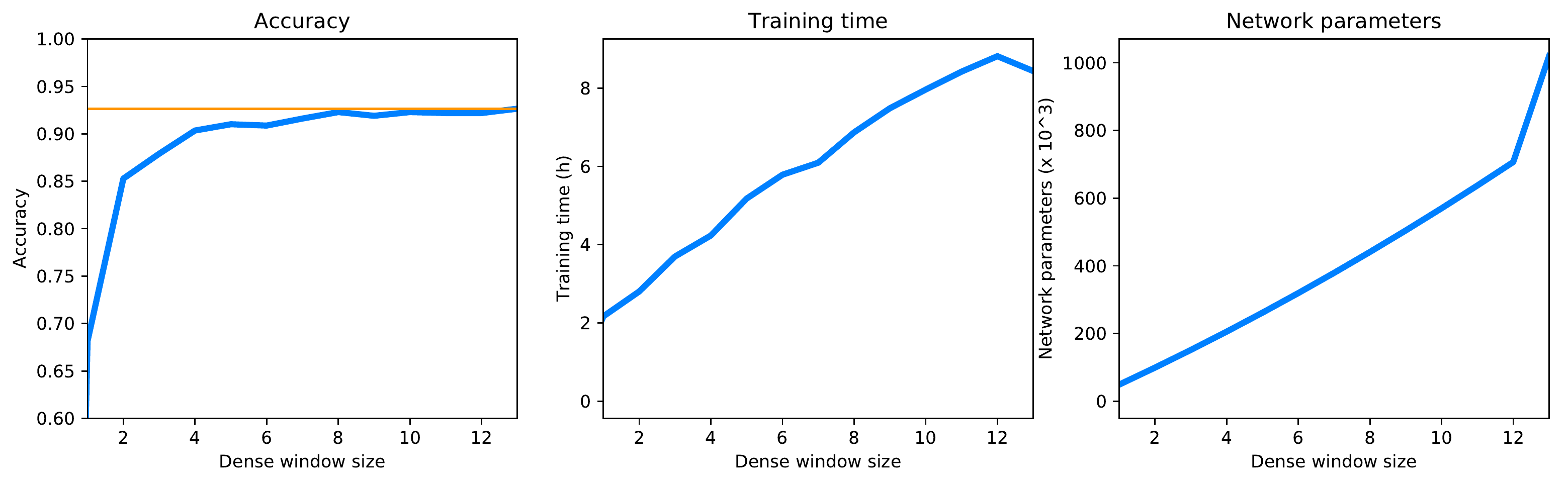}
   \caption{Accuracy, training time and number of trainable network parameters for densely connected networks at varying window sizes. For larger window sizes, note the large reductions in training time and number of parameters while accuracy is only slightly reduced. The orange line signifies the reference result of $92.65\%$ accuracy (at window size of $13$). The small dip in training time for window size $13$ is due to the simpler implementation of the straightforward DenseNet implementation in Tensorflow. The large parameter reduction going from a window size of $13$ to $12$ is due to transition (and the final) layers losing access to input maps that first entered the preceding block (there are $12$ convolutional layers in each dense block).}
   \label{figureMain}
  \end{center}
\end{figure}

The results in table \ref{tableResults} and figure \ref{figureMain} demonstrate that our proposed dense windowed connectivity pattern is able to greatly reduce the number of network parameters and training time with only modest reductions in accuracy for various window values. This provides evidence that the full connectivity pattern proposed in DenseNet [1] may not always be necessary in order to achieve good performance. With reduced training time one can expand hyper-parameter grid-search at train, and small models are more amenable to end-user applications (e.g. mobile), leading to the popularity of the \emph{network compression} field in recent years.

\begin{table*}
\medskip
\begin{adjustwidth}{-0.9in}{-1.0in}
\footnotesize
\begin{tabular}{| l | l | l | l |l | l | l | l |l | l | l | l |l | l | l | l | l | l |}
    \hline
    Window size & 1 & 2 & 3 & 4 & 5 & 6 & 7 & 8 & 9 & 10 & 11 & 12 & 13 \\ \hline
    Accuracy & 0.6815 & 0.8528 & 0.879 & 0.9035 & 0.9101 & 0.9087 & 0.9161 & 0.9229 & 0.919 & 0.9229 & 0.9219 & 0.9219 & 0.9265 \\ \hline
    Time (h) & 2.2 & 2.8 & 3.7 & 4.2 & 5.2 & 5.8 & 6.1 & 6.9 & 7.5 & 8.0 & 8.4 & 8.8 & 8.4 \\ \hline
    Parameters & 48882 & 99218 & 151450 & 205578 & 261602 & 319522 & 379338 & 441050 & 504658 & 570162 & 637562 & 706858 & 1019722 \\ \hline
    \end{tabular}
\end{adjustwidth}
\medskip
\caption{DenseNets with windowed dense connectivity train faster and have far fewer parameters with only a minor degradation in accuracy at larger dense window sizes.}
\label{tableResults}
\end{table*}

\subsection{Capacity Normalization}

In the previous section, our proposed method (WinDenseNet) was compared directly to DenseNet despite having far fewer trainable parameters (figure \ref{figureMain}). Here we wish to compare DenseNet-40 and WinDenseNet by normalizing for network capacity.

In order to normalize a network $A$ to have the same capacity as network $B$, we vary the growth rate of network A until reaching the same number of parameters as network B. However, since networks $A$ and $B$ have different structure, it is very unlikely that an integer value for growth rate will normalize to the exact capacity of $B$. For fair comparison, we therefore find the value $k$ such that $capacity(A(k)) < capacity(B) < capacity(A(k+1))$ and then train and obtain test accuracies for both $A(k)$ and $A(k+1)$. Finally, we linearly interpolate between the two capacity bounds in order to obtain a good estimate for the accuracy of a network with the same capacity as $B$.

\begin{figure}[t]
\begin{center}
   \includegraphics[width=1.0\textwidth]{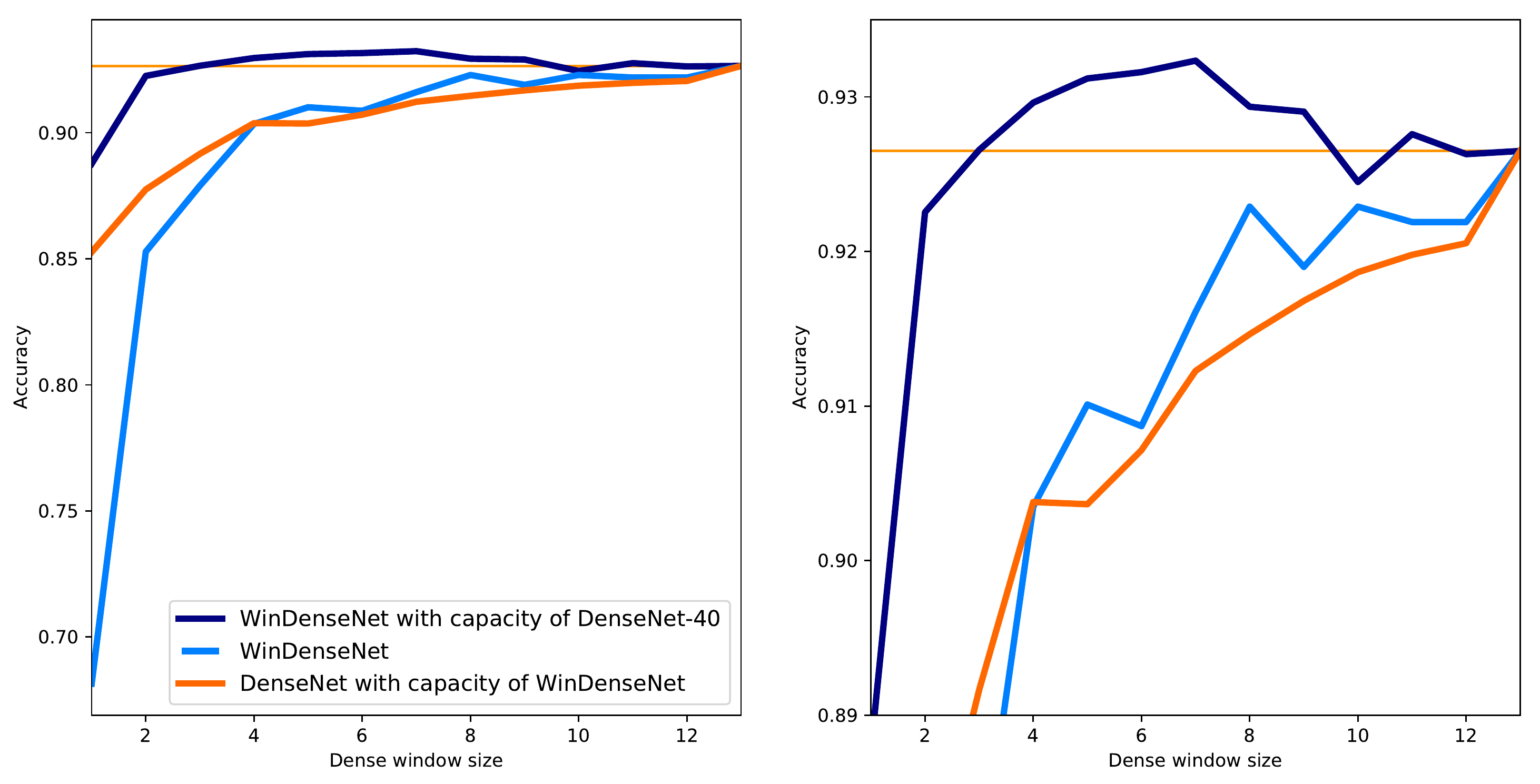}
   \caption{Our proposed method (WinDenseNet - light blue) demonstrates lower accuracy than DenseNet-40 (light orange horizontal line) albeit with far fewer parameters (see figure \ref{figureMain}). However, how would WinDenseNet perform if it had access to the same number of parameters as DenseNet-40? The answer is shown in dark blue; for certain values, WinDenseNet can utilize parameters more effectively. We also ask the question: how would DenseNet-40 perform if it only had access to the number of parameters found within WinDenseNet? The answer (in dark orange) is that for dense window sizes less than $4$, the fully dense connectivity of DenseNet performs better. However, for larger windows sizes WinDenseNet outperforms. Note that the right graph is a rescaled version of the graph on left. Networks are normalized by varying the growth rate (see text for more details).}
   \label{figureCapacityNormalized}
  \end{center}
\end{figure}

Having a normalization strategy in place, we capacity-normalize WinDenseNets at varying window sizes to each have the same capacity as the full DenseNet-40 (number of trainable parameters = $1,019,722$), and compare accuracy. The results are shown in dark blue in figure \ref{figureCapacityNormalized}. For certain dense window sizes, WinDenseNets not only benefit from the increased capacity but utilize this capacity more effectively than full DenseNet connectivity (light orange line). In other words, it can be more effective (say at window size $7$) to allocate network capacity toward increased growth rate \emph{rather than} allocate those same parameters toward increased dense connectivity.

Next, instead of increasing the capacity of WinDenseNets to match that of DenseNet-40, we decrease the capacity of DenseNet-40 to match the reduced capacity of WinDenseNet for each window size, and measure performance. These results are shown in dark orange in figure \ref{figureCapacityNormalized}. For small window sizes (less than $4$), and at their corresponding very low capacities (see Table \ref{tableResults}), the full connectivity provided by DenseNet prevails. However, for dense window sizes larger than $4$, figure \ref{figureCapacityNormalized} demonstrates that full dense connectivity can come at a cost; parameter resources allocated toward connectivity may, again,  be better applied toward increasing the number of feature maps at each layer instead (increasing growth rate).

\section{Discussion}

\begin{figure}
\centering
\begin{subfigure}{.5\textwidth}
  \centering
  \includegraphics[width=1.0\linewidth]{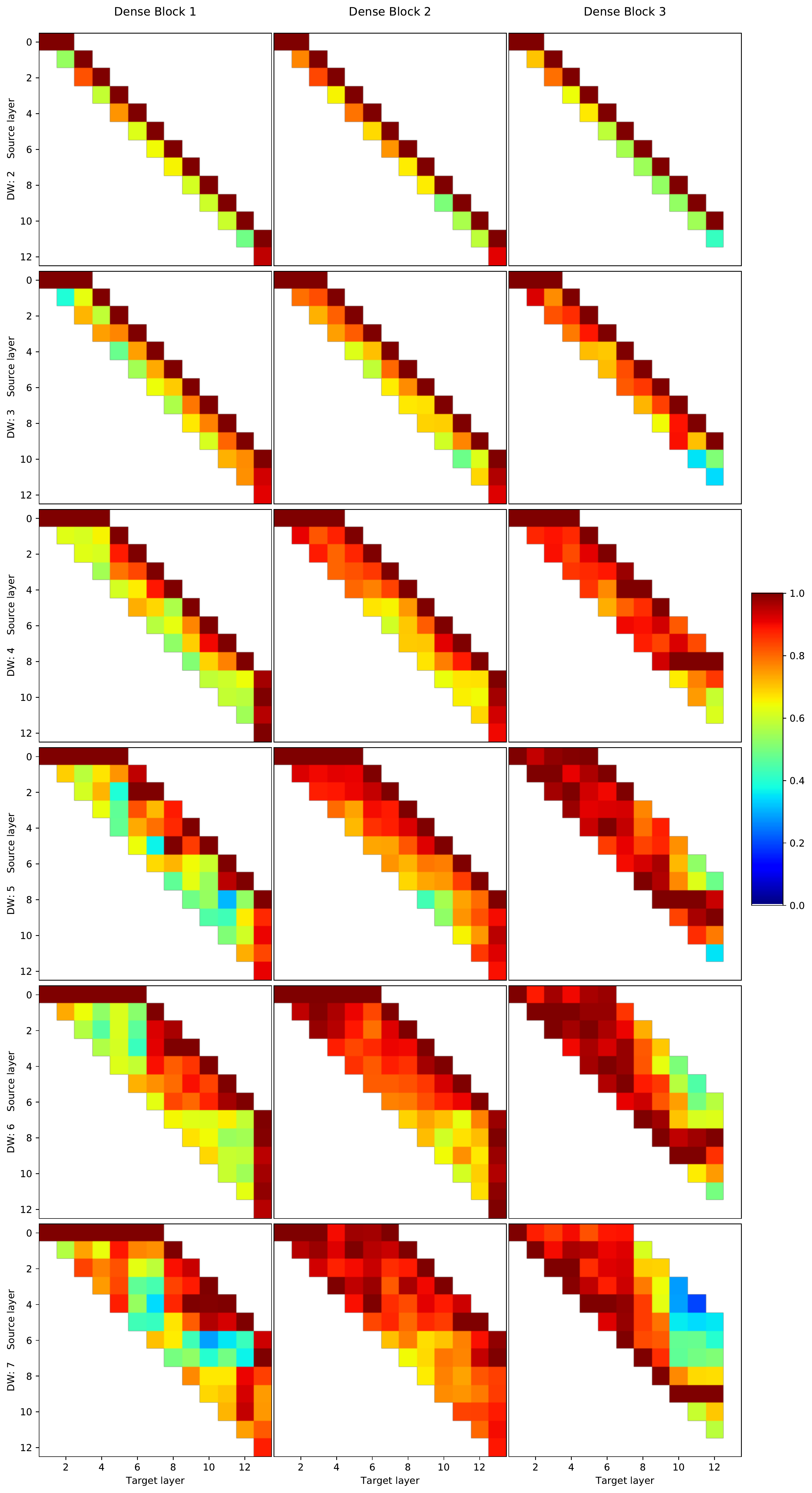}
  \label{fig:sub1}
\end{subfigure}%
\begin{subfigure}{.5\textwidth}
  \centering
  \includegraphics[width=1.0\linewidth]{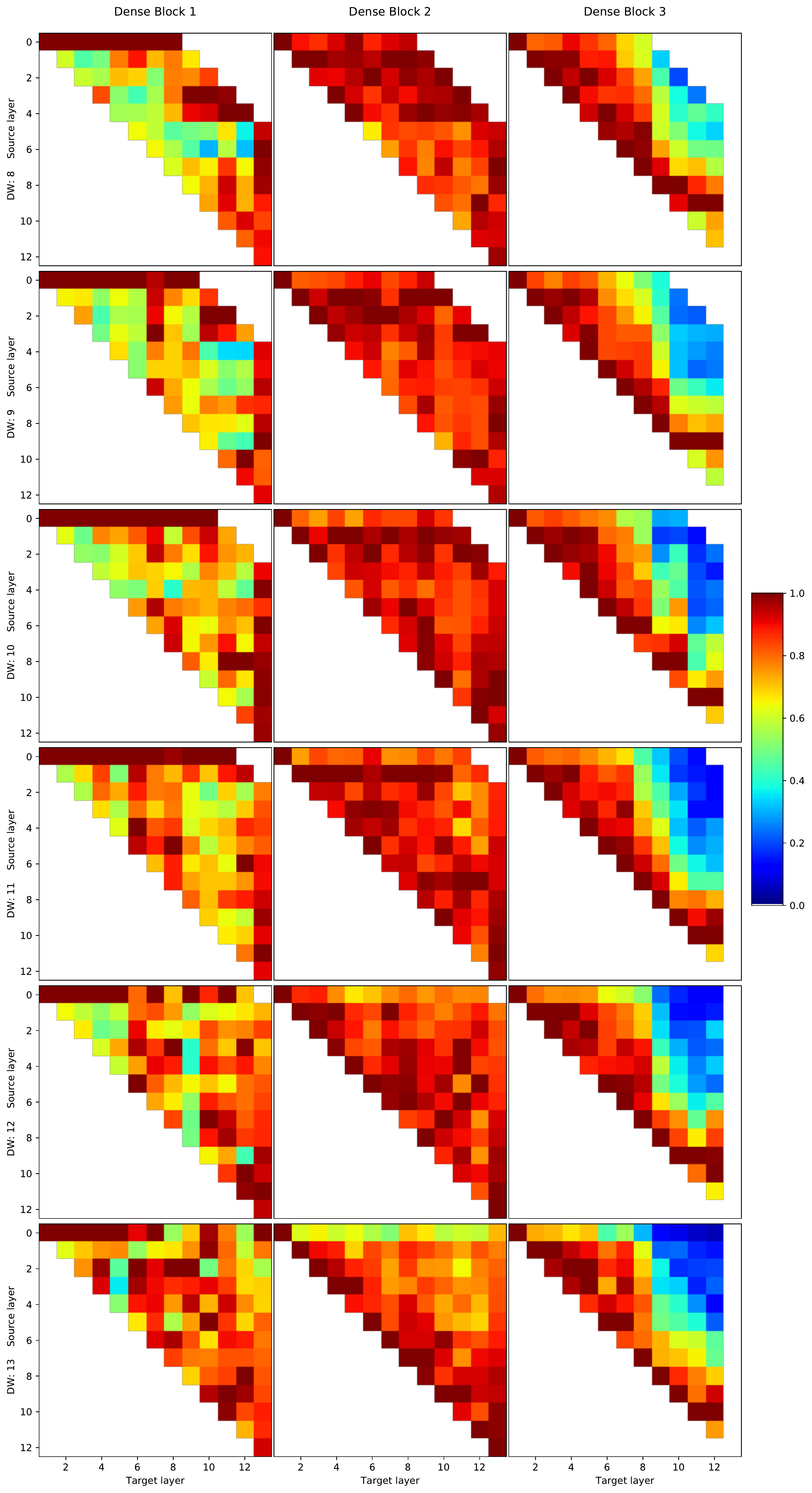}
  \label{fig:sub2}
\end{subfigure}
\caption{Dense block feature reuse in WinDenseNet architectures having dense window (DW) sizes between 2 and 13. Each column represents the relative strength of dependence on features from earlier layers (normalized to between 0 and max value). Note a DW value of 13 is equivalent to DenseNet [1]. WinDenseNets with small window sizes exhibit strong preference for the earliest features accessible while nets with larger window sizes prefer nearby features. See text for more details.}
\label{fig:figureFeatureReuse}
\end{figure}

One can measure the relative importance that a given target layer places on the feature maps its filters convolve over, by measuring the relative mean filter strengths corresponding to those source feature maps. This can provide some insight into how much a given layer \emph{reuses} features from previous layers. Given a target layer $L$ within a network having local dense connectivity of $w$, $L$ will be connected to between $1$ and $w$ preceding source layers. Each of the filters of layer $L$ convolve across the concatenation of feature maps from these source layers. Now we consider the mean value of all learned filter weights that correspond to feature maps from a given source layer which results in a single number \emph{for each} source layer. Finally, we normalize these mean filter strengths so that the maximum value is $1$ (each column is independently normalized so at least one value must be $1$). Therefore, each column in Figure \ref{fig:figureFeatureReuse} corresponds to a given target layer's \emph{relative} interest in (or reliance on) its input feature maps; in other words, how much a target layer \emph{reuses} feature maps from previous layers.

In figure \ref{fig:figureFeatureReuse}, one can see that networks with small dense window sizes have learned to benefit most from using feature maps from earlier layers (for each column, the highest value (dark red) occurs at the lowest source layer connection). This remains true within all three dense blocks.

At larger dense window sizes, networks begin to display the opposite affect: a tendency toward strongest feature reuse originating from the nearest source layers (for each column, the dark red value occurs at the bottom).

Another interesting observation is that, as the dense window size increases, feature reuse within each dense block becomes more diverse: \emph{dense block 1} exhibits more random feature reuse, \emph{dense block 2} exhibits a more consistent and strong reuse of prior features, and \emph{dense block 3} shows diminishing interest in features from more distance source layers.

\begin{figure}[t]
\begin{center}
   \includegraphics[width=0.9\textwidth]{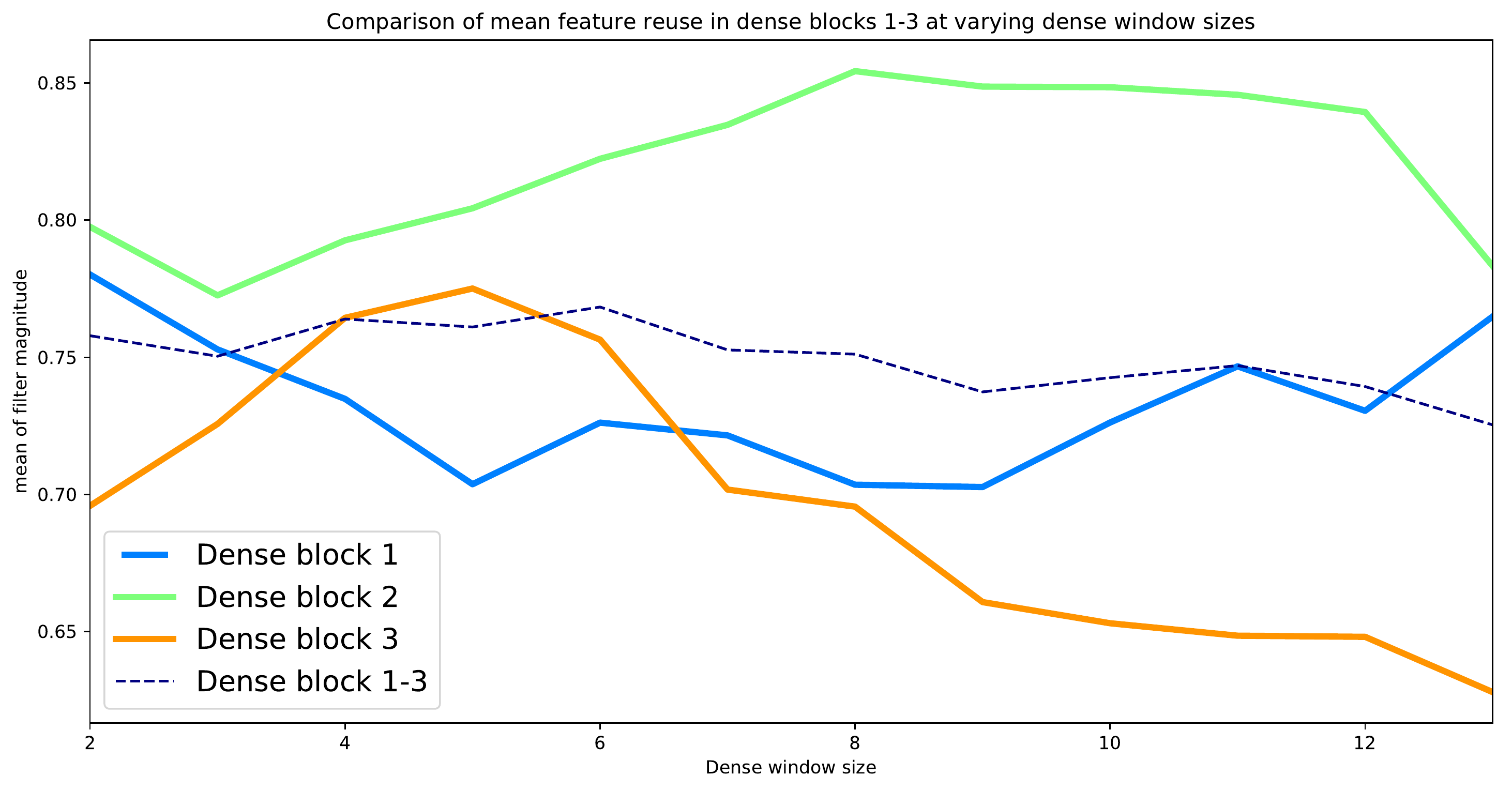}
   \caption{Mean of normalized filter values found within each dense block for networks with varying dense connectivity window sizes (mean of colored values found in each block of figure \ref{fig:figureFeatureReuse}). The dashed line is the mean across all three dense blocks. In block 3, as well as the mean across all three blocks, one can see a trend toward decreased feature reuse at larger dense window sizes.}
   \label{figureFeatureReuseForVaryingDW}
  \end{center}
\end{figure}

Also, one can see in figure \ref{fig:figureFeatureReuse} that different dense blocks reuse features that first enter each dense block to varying degrees (top rows correspond to maps entering dense blocks). \emph{Dense block 1} exhibits strong interest in this input layer at almost all dense window sizes. \emph{Dense block 2} has a strong interest for small window sizes and lower interest for large window sizes (\emph{Dense block 3} even less so).

This analysis provides some evidence why it may be unnecessary to densely connect networks to the fullest extent as in [1]; if a network can learn to achieve good performance by only reusing features in a local window, then network capacity allocated toward further connectivity would be better applied to added representational expression (more filters/feature maps). This can be seen especially in \emph{Dense block 3} and this provides some justification for our results in figure \ref{figureCapacityNormalized}; where mid-sized dense connectivity networks led to the highest accuracy when capacity-normalized. 

Lastly, figure \ref{figureFeatureReuseForVaryingDW} displays the normalized mean filter strength within each dense block for varying dense window sizes (the mean of colored values in each block shown in figure \ref{fig:figureFeatureReuse}). Once again one can see a declining trend away from feature reuse of earlier layers especially for \emph{Dense block 3} but also within the mean of all dense blocks (dashed line).

\section{Conclusion/Future Work }
The success of full dense connectivity [1] rests upon the idea that reusing features from previous layers can be more important than adding new features at each layer (full connectivity with low growth rates). In this work, by introducing the notion of \emph{local dense connectivity}, we have shown that there is indeed a trade-off between the amount of dense connectivity and the amount of representational expression available at each layer (\# of filters/feature maps). In other words, the fully-dense connectivity pattern of DenseNet may not always be necessary and network parameter resources may be better put toward \emph{a combination of local dense connectivity combined with increased growth rate}. These findings were further supported by an analysis of \emph{to what extent} features were being reused at various layers. In section 4.3 and figure 3 of [1], the authors show that DenseNets can make more efficient use of parameters than ResNets [4] and our proposed local dense connectivity pattern builds upon this showing even further parameter efficiency gains are possible.

Our examination of feature reuse provides some evidence that different dense blocks could benefit from having different amounts of dense connectivity - an interesting avenue for future work. As well, it may be fruitful to explore the interaction between locally dense networks and bottleneck and transition layer compression as part of an expanded study of the other networks found in [1]. The potential parameter efficiency of locally dense networks should also be useful for other tasks that utilize dense networks such as semantic segmentation [13].

\section*{References}

\medskip
\small
[1] G. Huang, Z. Liu, L. van der Maaten, and K. Weinberger. “Densely Connected Convolutional Networks,” {\it 2017 IEEE Conference on Computer Vision and Pattern Recognition (CVPR)}, 2017.

[2] A. Krizhevsky, I. Sutskever, and G. E. Hinton. "Imagenet classification with deep convolutional neural networks," {\it Neural Information Processing Systems (NIPS)}, 2012.

[3] K. Simonyan, A. Zisserman. "Very deep convolutional networks for large-scale image recognition," {\it arXiv preprint arXiv:1409.1556}, 2014.

[4] K. He, X. Zhang, S. Ren, and J. Sun. "Deep residual learning for image recognition," {\it 2016 IEEE Conference on Computer Vision and Pattern Recognition (CVPR)}, 2016.

[5] S. Hochreiter, and J. Schmidhuber. "Long short  term  memory," {\it Technical  Report  FKI-207-95, Technische Universitaet Muenchen}, 1995.

[6] R. K. Srivastava, K. Greff, and J. Schmidhuber. "Training very deep networks," {\it Neural Information Processing Systems (NIPS)}, 2015.

[7] G. Huang, Y. Sun, Z. Liu, D. Sedra, K., and Q. Weinberger. "Deep Networks with Stochastic Depth," {\it European Conference on Computer Vision}, 2016.

[8] G. Larsson, M. Maire, and G. Shakhnarovich. "Fractalnet: Ultra-deep neural networks without residuals," {\it arXiv preprint arXiv:1605.07648}, 2016

[9] Tensorflow - https://www.tensorflow.org

[10] https://github.com/LaurentMazare/deep-models/tree/master/densenet

[11] https://github.com/liuzhuang13/DenseNet

[12] A. Krizhevsky, G. Hinton, "Learning multiple layers of features from tiny images," {\it Tech Report}, 2009


[13] S. Jégou, M. Drozdzal, D. Vazquez, A. Romero, and Y. Bengio. "The one hundred layers tiramisu: Fully convolutional densenets for semantic segmentation," {\it 2017 IEEE Conference on Computer Vision and Pattern Recognition Workshops (CVPRW)}, 2017.


\end{document}